\documentclass{article}
\usepackage{CJKutf8}

\usepackage{arxiv}

\usepackage[utf8]{inputenc} % allow utf-8 input
\usepackage[T1]{fontenc}    % use 8-bit T1 fonts
\usepackage{hyperref}       % hyperlinks
\usepackage{url}            % simple URL typesetting
\usepackage{booktabs}       % professional-quality tables
\usepackage{amsfonts}       % blackboard math symbols
\usepackage{nicefrac}       % compact symbols for 1/2, etc.
\usepackage{microtype}      % microtypography
\usepackage{lipsum}
\usepackage{graphicx}
\usepackage{color}
\usepackage{multirow}
\usepackage{tablefootnote}

\usepackage{xcolor}
\usepackage{listings}
\graphicspath{ {./images/} }

\title{Syntactic-GCN Bert based Chinese Event Extraction}

\author{
 Jiangwei Liu \\
  School of Information Management and Engineering\\
  Shanghai University of Finance and Economics\\
  Shanghai 200433, China \\
  \texttt{majorliujw@gmail.com} \\
   \And
 Jingshu Zhang \\
  School of Information Management and Engineering\\
  Shanghai University of Finance and Economics\\
  Shanghai 200433, China \\
  \texttt{cj19660606@163.sufe.edu.cn} \\
  \And
 Xiaohong Huang \\
  School of Information Management and Engineering\\
  Shanghai University of Finance and Economics\\
  Shanghai 200433, China \\
  \texttt{huangxiaohong@163.sufe.edu.cn} \\
    \And
  Liangyu Min\thanks{Corresponding author.} \\
  School of Information Management and Engineering\\
  Shanghai University of Finance and Economics\\
  Shanghai 200433, China \\
  \texttt{minux@163.sufe.edu.cn} \\

}

\begin{document}
\maketitle
\begin{abstract}
With the rapid development of information technology, online platforms (e.g., news portals and social media) generate enormous web information every moment. Therefore, it is crucial to extract structured representations of events from social streams. Generally, existing event extraction research utilizes pattern matching, machine learning, or deep learning methods to perform event extraction tasks. However, the performance of Chinese event extraction is not as good as English due to the unique characteristics of the Chinese language. In this paper, we propose an integrated framework to perform Chinese event extraction. The proposed approach is a multiple channel input neural framework that integrates semantic features and syntactic features. The semantic features are captured by BERT architecture. The Part of Speech (POS) features and Dependency Parsing (DP) features are captured by profiling embeddings and Graph Convolutional Network (GCN), respectively. We also evaluate our model on a real-world dataset. Experimental results show that the proposed method outperforms the benchmark approaches significantly.

\end{abstract}

% keywords can be removed
%\keywords{First keyword \and Second keyword \and More}

\section{Introduction}
Online platforms generate an enormous quantity of web information every day. For example, news portals report real-time news; social media broadcast the hot news and generate related topics every moment. Information Extraction (IE) has gained increasing popularity because it helps exploit this potential by automatically extracting content from massive information. \cite{DoddingtonMitchell-1}. As a particular form of Information Extraction (IE), Event Extraction (EE) has gained increasing popularity due to its ability to automatically extract events from human language \cite{LiuMin-150}. Event Extraction initially started in the late 1980s when the U.S. Defense Advanced Research Projects Agency (DARPA) boosted research into message understanding \cite{HogenboomFrasincar-2}. Now event extraction has become an important and challenging task, which aims to deal with the "5W1H" questions, i.e., "who", "when", "where", "what", "why" and "how" of an event. Event extraction is closely related to computer science, statistics, and natural language processing as an interdisciplinary subject. As a particular form of information, event extraction involves named entity recognition (NER) and relation extraction (RE), and mostly depends on the results of these tasks. 

\begin{figure}[htbp]
	\centering
	\includegraphics[width= 6 in]{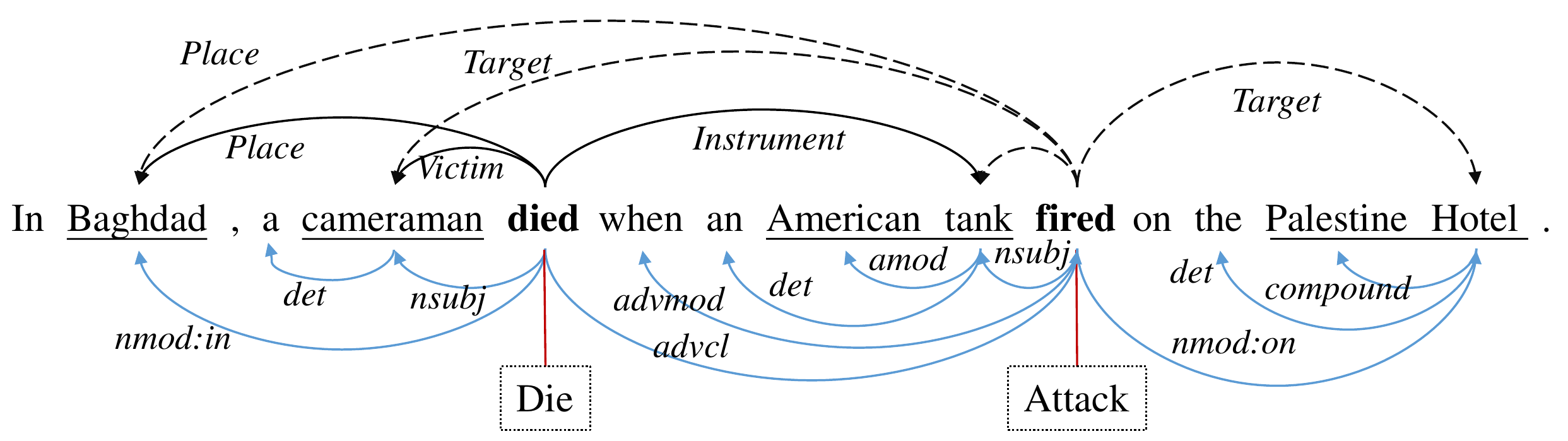}	
	\caption{An example of two events in one sentence: Die and Attack. The upper arcs link event triggers to their corresponding arguments, with the argument roles on the arcs. The lower side demonstrates the syntactic parser results.}
	\label{fig1}	
\end{figure}

\begin{figure}[htbp]
	\centering
	\includegraphics[width= 5.5 in]{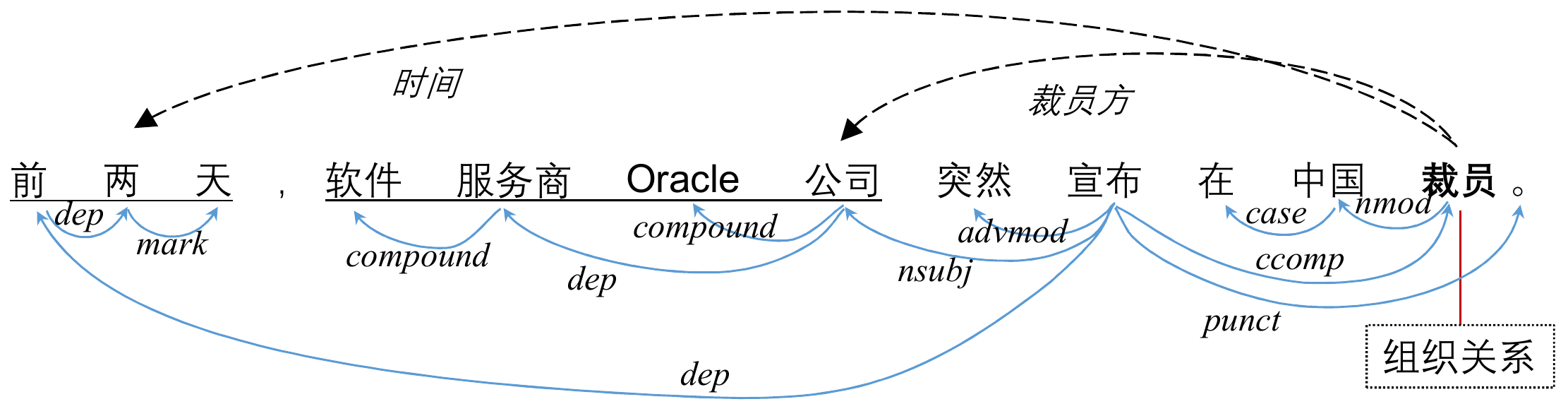}	
	\caption{An example of Chinese Event: event type is "\begin{CJK*}{UTF8}{gbsn}组织关系 - 裁员\end{CJK*}". The upper arcs link event triggers to their corresponding arguments, with the argument roles on the arcs. The lower side demonstrates the syntactic parser results.}
	\label{fig2}	
\end{figure}

Following the event extraction task definition in ACE 2005, an event is frequently described as a change of state, indicating a specific occurrence of something that happens in a particular time and a specific place involving one or more participants. We give an English example in Fig. \ref{fig1} and a Chinese example in Fig. \ref{fig2}, respectively. For example, there are two event types involved in sentence S1: "Die" and "Attack", triggered by "died" and "fired", respectively. For Die event, "Baghdad", "cameraman", and "American tank" are its arguments with corresponding roles: Place, Victim, and Instrument, respectively. For Attach event, "Baghdad", "cameraman", "American tank" and "Palestine Hotel" are its arguments with corresponding roles: Place, Victim, Instrument and Target, respectively. This is a somewhat more complex example with three arguments shared, which is more challenging than the simple case with one event type in one sentence. Figure \ref{fig1} shows the English event extraction annotation and the syntactic parser results. In Fig. \ref{fig2}, the event type is "\begin{CJK*}{UTF8}{gbsn}组织关系\end{CJK*}" and the sub event type is "\begin{CJK*}{UTF8}{gbsn}裁员\end{CJK*}" triggered by "\begin{CJK*}{UTF8}{gbsn}裁员\end{CJK*}". "\begin{CJK*}{UTF8}{gbsn}前两天\end{CJK*}" and  "\begin{CJK*}{UTF8}{gbsn}软件服务商Oracle公司\end{CJK*}" are event arguments with corresponding roles: "\begin{CJK*}{UTF8}{gbsn}时间\end{CJK*}" and "\begin{CJK*}{UTF8}{gbsn}裁员方\end{CJK*}".

Due to its outstanding usefulness, event extraction has been applied in various fields, such as the biomedical \cite{YakushijiTateisi-42, KilicogluBergler-43, BuykoFaessler-15}, general information extraction \cite{YangarberGrishman-44, LeeChen-45}, finance and economics \cite{BorsjeHogenboom-46, LiuHuang-151}, news recommendation \cite{HanHuang-152}, hate crime prediction \cite{HanHuang-153} etc.

Event extraction can be divided into closed-domain and open-domain event extraction. From the view of techniques used, existing approaches can be divided into four categories: pattern matching \cite{YangarberGrishman-44, LeeChen-45}, machine learning \cite{HennSticha-16, PengMoh-13, LuRoth-60}, deep learning \cite{LuLin-65, AhmadPeng-66, ZhaoZhang-67, LybargerOstendorf-68, LybargerOstendorf-69, MinRozonoyer-70, ShengGuo-17, CaselliMutlu-71, MaoLi-72}, and semi-supervised learning methods \cite{HuangJi-84, MansouriNaderan-Tahan-85, MiyazakiKomatsu-86, ZhouChen-87, ChenWang-88}. Feature engineering is the main challenging issue of traditional event extraction methods. And traditional machine learning methods have limitations in learning deep or complex nonlinear relations. Deep learning methods can alleviate these shortages and thus have become the mainstream event extraction methods.

Deep learning methods can learn distributed representation of knowledge, e.g., semantic features, avoiding feature engineering. Word embedding, character embedding, position embedding, entity type embedding, POS tag embedding, entity type embedding, word distance, relative position, path embedding, etc., are the most used features \cite{ZhaoZhang-67, BjorneSalakoski-78, LiuLuo-77}. Except for the multi-channel distributed representation of the input, researchers have employed some techniques to capture the features contained in these representations. For example, to better capture the complex relationships among local and global contexts in biomedical documents, Zhao et al. \cite{ZhaoZhang-67} use a dependency-based GCN network to capture the local context and a hypergraph to model the global context. In addition, the fine-grained interaction between the local and global contexts is captured by a series of stacked Hypergraph Aggregation Neural Network (HANN) layers. Inspired by \cite{ZhaoZhang-67}, we propose an integrated approach that considers semantic features and syntactic features to extract Chinese events. In our framework, the semantic features are captured by BERT \cite{DevlinChang-154}. The Part of Speech (POS) features are initialed by random variables matrix. The Dependency parsing (DP) features are obtained by Stanford NLP \footnote[1]{https://nlp.stanford.edu/} and transformed into a Graph Convolutional Network (GCN) from its parsing tree. All the POS and DP embeddings are treated as tensor variables and trained during the backpropagation. The difference between our approach and \cite{ZhaoZhang-67} is that \cite{ZhaoZhang-67} adopted sequence input: word embedding, position embedding, and entity type embedding are treated as input of stacked GCN layers. While our work embraces multiple channel input: semantic features generated by BERT, POS embeddings, and DP GCN embeddings are concatenated as input of fully connected layers.

We summarize the contributions of this study as follows: 

(1) We propose an integrated framework to perform Chinese Event Extraction. The framework integrates the semantic features (BERT) and syntactic features (Part of Speech and Dependency Parsing). Considering the unique characteristics of the Chinese language, we adopt alignment of POS embeddings and DP embeddings to BERT input tokens.

(2) We also verify the proposed approach on an empirical dataset. Empirical experiments show that the proposed method can significantly enhance the extraction performance compared with BERT pre-trained models.

The remainder of this paper is organized as follows. We introduce the empirical dataset and its preprocessing in section 2 and propose our framework in section 3. Section 4 reports experimental results and discusses the reasonability, followed by the conclusion in Section 5.

\section{Dataset and Preprocessing}

\subsection{Dataset}
We choose the Chinese Event Extraction Dataset (DuEE 1.0) adopted in Language and Intelligent Technology Competition 2020\footnote{https://aistudio.baidu.com/aistudio/competition/detail/32/0/introduction}  and 2021\footnote{https://aistudio.baidu.com/aistudio/competition/detail/65/0/introduction}. DuEE1.0 corpus is selected from the hot search board of Baidu, which reflects various interests of most Chinese people. The dataset includes a list of 65 pre-defined event types, a training set (12000 sentences), a development set (1500 sentences) and two test sets (3500 sentences), containing 17000 sentences total.

Considering that the 3500 labeled test sets are not public to users, we divide the 1500 development set into a new validation set (500 sentences) and a new test set (1000 sentences). In the following experiments, we train our model on the 12000 training set, choose the best model by the new validation set (500 sentences), and test the results on the new test set (1000 sentences).

\begin{figure}[htbp]
	\centering
	\includegraphics[width= 6.5 in]{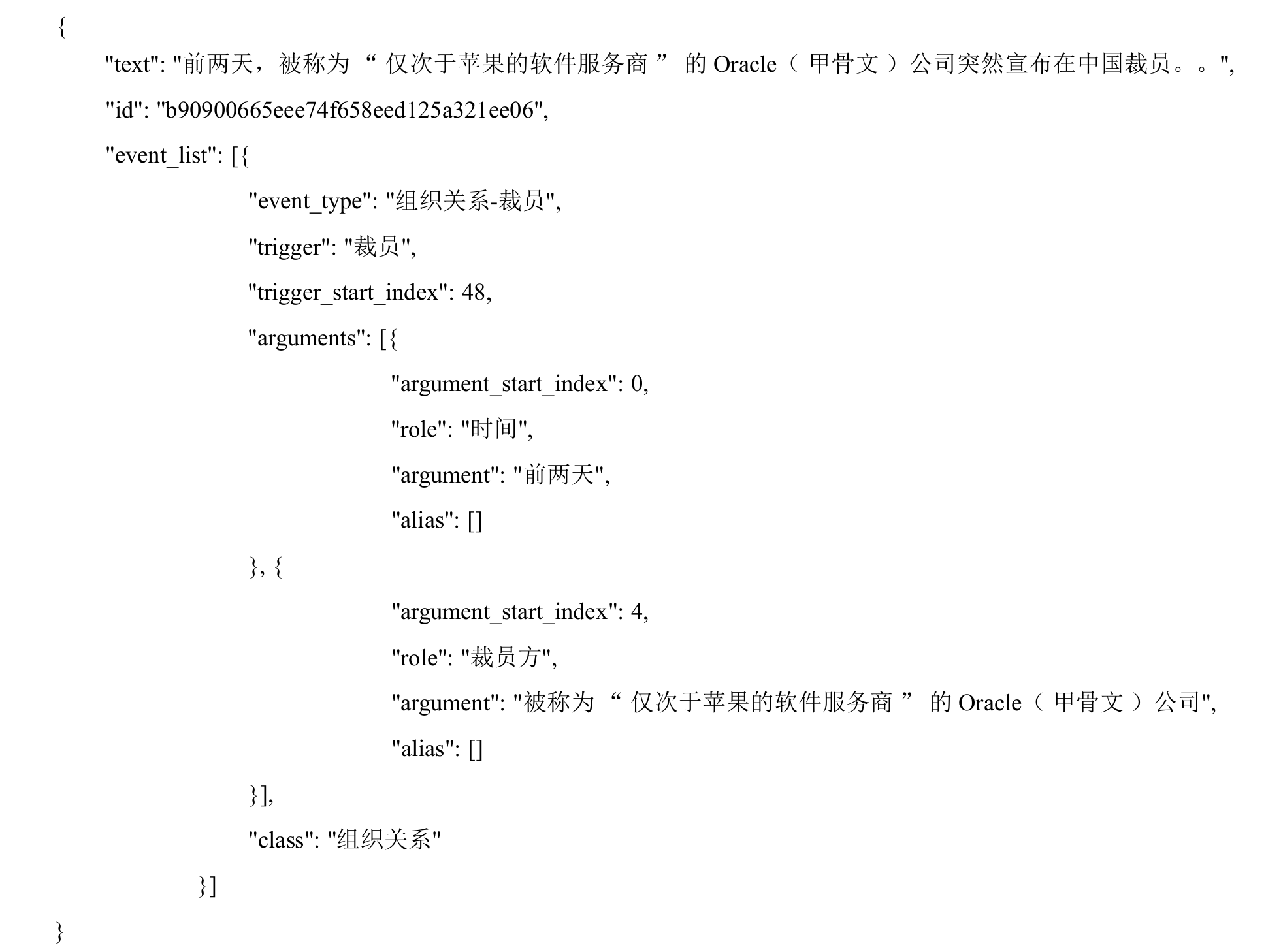}	
	\caption{One annotated example from the Chinese Event Extraction Dataset (DuEE 1.0).}
	\label{fig3}	
\end{figure}

One annotated example is shown Fig. \ref{fig3}.

\subsection{Preprocessing}
Some preprocessing needs to be performed before carrying out the experiments.

Firstly, resplit the new validation and test sets as described in the previous section.

Secondly, perform NER, Part of Speech (POS), Dependency Parsing (DP) using Stanford NLP.

Thirdly, align the POS, NER, and DP tags with BERT input tokens.

Lastly, reorganize the datasets.

\section{Our Approach}
\begin{figure}[htbp]
	\centering
	\includegraphics[width= 6.5 in]{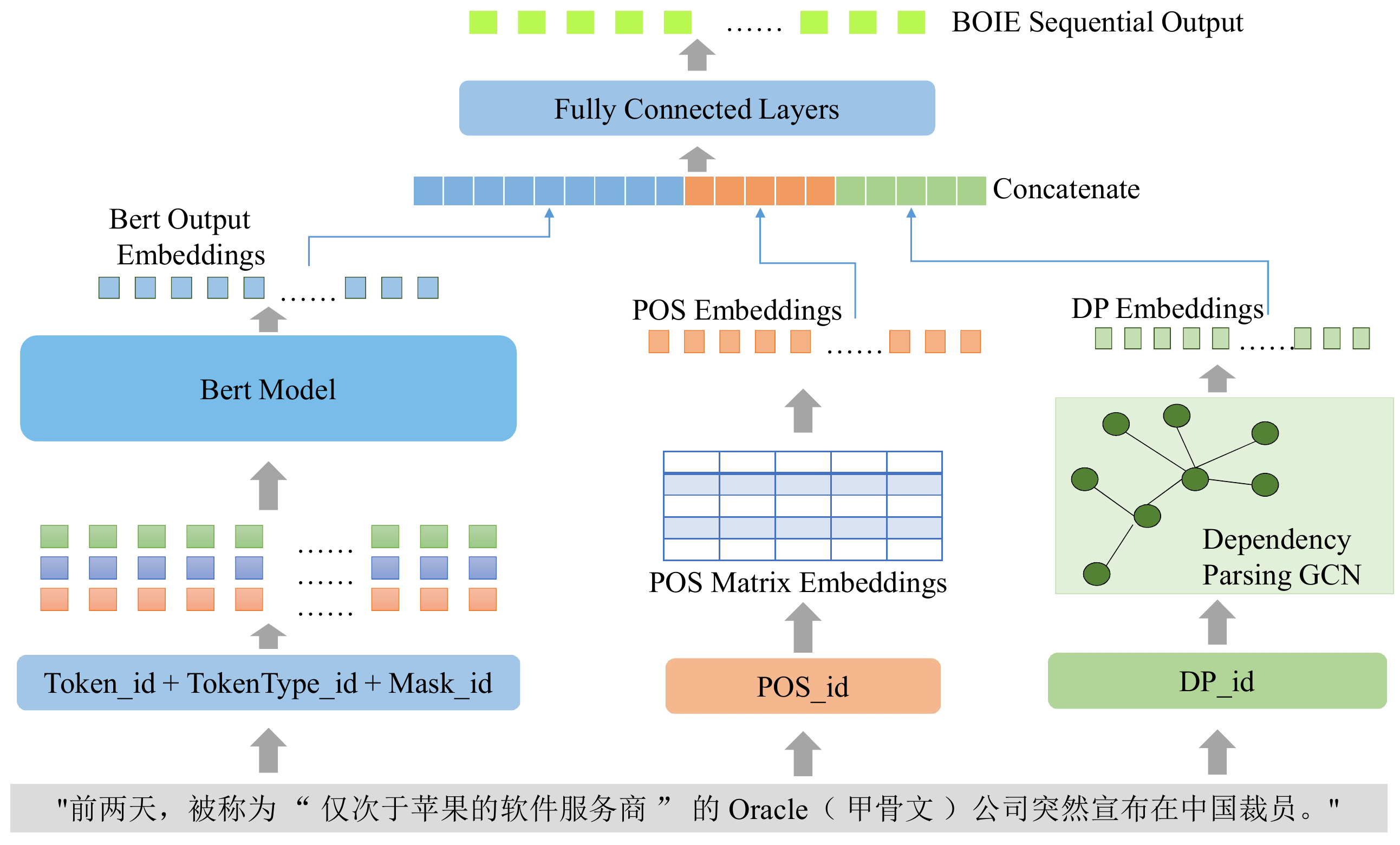}	
	\caption{Different Event Extraction classification.}
	\label{fig4}	
\end{figure}

Our framework contains three main modules: BERT module, Part of Speech embedding module, and Dependency Parsing GCN embedding module. The architecture is demonstrated in Fig. \ref{fig4}.

\section{Experiment}
In this section, we mainly report the experimental results. Firstly, the event extraction evaluation metrics are introduced before demonstrating the experimental results. Then, we report the experimental results and provide analyses of the observations. Lastly, we discuss the effectiveness of the syntactic features and the shortcomings of our approach. We also list the planned experiments in the future.

\subsection{Event Extraction Evaluation Metrics}

The event extraction task, especially the closed-domain event extraction task, can be regarded as a classification task or Sequence Labeling task. Most existing literature uses classification metrics to evaluate the event extraction performance. In accordance with IE and TM, performance is generally measured by calculating the quantity of true positives and negatives, as well as that of false positives and negatives. The most used metrics, e.g., precision, recall, and F1 score, are calculated as follows:

\begin{equation} 
	Precision = \frac{{TP}}{{TP + FP}}
\end{equation}
\begin{equation} 
	Recall = \frac{{TP}}{{TP + FN}}
\end{equation}
\begin{equation} 
	F1 = \frac{{2*Precision*Recall}}{{Precision + Recall}} = \frac{{2*TP}}{{2*TP + FP + FN}}
\end{equation}

These performance measures provide a brief explanation of the "Confusion Metrics". True positives (TP) and true negatives (TN) are the observations that are correctly predicted. In contrast, false positives (FP) and false negatives (FN) are the values that the actual class contradicts with the predicted class.

Following the DUEE competition evaluation rules, evaluation is case insensitive and calculated according to the token level. If an event argument has multiple annotated mentions, the one with the highest matching F1 will be used.

\subsection{Benchmark and Our Approach Variant Models}
We choose the state-of-the-art pre-trained model BERT as a benchmark in our experiment. We also provide variants of our method. Benchmark models and our approach variants are listed as follows:

\begin{itemize}
	\item  \textbf{BERT.} BERT is a bi-directional transformer architecture model, which has been trained on massive corpora and has learned fairly good semantic representations conditioned on token context and remains rich textual information. Recently, much research has used Bert pre-trained representation as shared textual input features of Event Extraction models. In our experiments, we use two variant BERT Models (e.g., RTB3 and RoBERTa-wwm-ext).
	
	\item  \textbf{POS-DP-LSTM-BERT.} The input channels contain Part of Speech embeddings (POS), Dependency Parsing embeddings (DP), and semantic embeddings (BERT). One LSTM model the POS sequence, and another LSTM model the DP sequence. Finally, the fully connected layers model the concatenated embeddings (the two LSTM output and BERT output embeddings) and predict the BIOE labels.
	
	\item  \textbf{POS-Embedding-BERT.} The input channels contain Part of Speech embeddings (POS) and semantic embeddings (BERT). The POS embeddings are obtained by looking up the POS embedding matrix by POS id. Finally, the fully connected layers model the concatenated embeddings (the POS embeddings and BERT output embeddings) and predict the BIOE labels.
	
	\item  \textbf{DP-GCN-BERT.} The input channels contain Dependency Parsing embeddings (DP) and semantic embeddings (BERT). The DP embeddings are obtained by looking up the Dependency Parsing Graph Convolutional Network by DP id. Finally, the fully connected layers model the concatenated embeddings (the DP embeddings and BERT output embeddings) and predict the BIOE labels.
	
	\item  \textbf{POS-DP-GCN-BERT.} This model integrated the POS-Embedding-BERT model and DP-GCN-BERT model. The input channels contain Part of Speech embeddings (POS), Dependency Parsing embeddings (DP), and semantic embeddings (BERT). The POS embeddings are obtained by looking up the POS embedding matrix by POS id. The DP embeddings are obtained by looking up the Dependency Parsing Graph Convolutional Network by DP id. Finally, the fully connected layers model the concatenated embeddings (the POS embeddings, DP embeddings, and BERT output embeddings) and predict the BIOE labels.

\end{itemize}

\subsection{Experiment results}
The overall event extraction results are shown in Table \ref{tab2}. RBT3 and RoBERTa-wwm-ext are two different variants of Chinese BERT pre-trained models. The most significant difference is the number of hidden layers: RBT3 has three hidden layers while RoBERTa-wwm-ext has twelve hidden layers.
\begin{table}
	\caption{The overal event extraction results.}
	\centering
	\begin{tabular} {p{70pt} p{110pt} p{45pt} p{45pt} p{45pt} p{50pt}} %{llllll} 
		\toprule %\hline
		BERT Models & Models	& Precision	& Recall	& F1 & Loss/epoch \\
		\midrule %\hline
		\multirow[c]{4}{0.6 in}{RBT3}	& BERT & 0.6064 	& 0.7167 	& 0.6555 	& 0.2170	\\
		& POS-DP-LSTM-BERT	& 0.6182 	& 0.7368 	& 0.6700 	& 0.2692 	\\
		& POS-Embedding-BERT	& 0.9013 	& 0.9646 	& 0.9307 	& 0.0255 	\\	
		& DP-GCN-BERT	& 0.8995 	& 0.9605 	& 0.9280 	& 0.0236 	\\	
		& POS-DP-GCN-BERT	& 0.8886 	& 0.9565 	& 0.9200 	& 0.0274 	\\
		\midrule %\hline
		\multirow[c]{4}{0.6 in}{RoBERTa-wwm-ext}	& BERT & 0.6703 	& 0.7525 	& 0.7082 	& 0.2009 	\\
		& POS-DP-LSTM-BER	& 0.6545 	& 0.7348 	& 0.6912 	& 0.2472 	\\
		& POS-Embedding-BERT	& 0.8946 	& 0.9240 	& 0.9088 	& 0.0297 	\\	
		& DP-GCN-BERT	& 0.8889 	& 0.9157 	& 0.9020 	& 0.0294 	\\	
		& POS-DP-GCN-BERT	& 0.8925 	& 0.9290 	& 0.9101 	& 0.0278 	\\
		
		\bottomrule %\hline
	\end{tabular}
	\label{tab2}
\end{table}

From Table \ref{tab2}, some findings can be summarized as follows:

Firstly, considering the two benchmark models, RoBERTa-wwm-ext BERT model (70.82\%) surpasses RBT3 BERT model (65.55\%).

Secondly, part of speech features (POS Embeddings) can significantly improve the event extraction performance, no matter which BERT pre-trained model is used as a benchmark model. For example, POS-Embedding-BERT(RBT3) obtains 93.07\% F1 score, and POS-Embedding-BERT(RoBERTa-wwm-ext) obtains 90.88\& F1 score.

Thirdly, dependency parsing features (DP GCN Embeddings)  can significantly improve the event extraction performance, no matter which BERT pre-trained model is used as a benchmark model. For example, DP-GCN-BERT(RBT3) obtains 92.80\% F1 score, and DP-GCN-BERT(RoBERTa-wwm-ext) obtains 90.20\& F1 score.

Fourthly, We integrate part of speech features and dependency parsing features to check whether they can further enhance the performance. When we choose the RoBERTa-wwm-ext BERT model as the benchmark, we find that the results are what we expected. The integrated model POS-DP-GCN-BERT achieves the best performance (91.01\% F1 score). However, when the RBT3 is selected as the benchmark model, we find that the integrated model POS-DP-GCN-BERT (92.00\% F1 score) is inferior to POS-Embedding-BERT (93.07\% F1 score) and DP-GCN-BERT (92.80\% F1 score), which only combine one kind of syntactic feature. We will explore the reasons in the discussion section.

Finally, we attribute the performance boost to multiple channel input information. BERT offers semantic features, while Part of Speech (POS) and Dependency Parsing (DP) provide syntactic features. They are not conflicted but complementary.

\subsection{Discussion}
The closed-domain event extraction task can be divided into four subtasks: trigger identification, event type classification, argument identification, and argument role classification. From the manner of how to organize the subtasks of the event extraction, most of the existing closed-domain event extraction methods can be divided into two mainstreaming categories: pipelined-based method and joint-based method. To demonstrate the effectiveness of the Part of Speech and Dependency Parsing, we use the pipelined manner to train the models. We report the extraction results of the trigger identification and role classification subtasks in Table \ref{tab3} and Table \ref{tab4}.

\begin{table}
	\caption{The trigger identification and role classification results based on pre-trained RBT3}
	\centering
	\begin{tabular} {p{70pt} p{110pt} p{45pt} p{45pt} p{45pt} p{50pt}} %{llllll} 
		\toprule %\hline
		Tasks	& Models	& Precision	& Recall	& F1 & Loss/epoch \\
		\midrule %\hline
		\multirow[c]{4}{0.6 in}{Trigger}	& BERT & 0.7654	& 0.8372	& 0.7997	& 0.0727	\\
		& POS-DP-LSTM-BERT	& 0.778	& 0.8414	& 0.8085	& 0.0565	\\
		& POS-Embedding-BERT	& 0.9934	& 0.9958	& 0.9946	& 0.0009	\\	
		& DP-GCN-BERT	& 0.9901	& 0.995	& 0.9925	& 0.0005	\\	
		& POS-DP-GCN-BERT	& 0.9934	& 0.9983	& 0.9959	& 0.0007	\\
		\midrule %\hline
		\multirow[c]{4}{0.6 in}{Role}	& BERT & 0.4474	& 0.5962	& 0.5112	& 0.3613	\\
		& POS-DP-LSTM-BER	& 0.4583	& 0.6322	& 0.5314	& 0.4819	\\
		& POS-Embedding-BERT	& 0.8091	& 0.9334	& 0.8668	& 0.0500		\\	
		& DP-GCN-BERT	& 0.8088	& 0.9259	& 0.8634	& 0.0467		\\	
		& POS-DP-GCN-BERT	& 0.7837	& 0.9146	& 0.8441	& 0.0540		\\
		
		\bottomrule %\hline
	\end{tabular}
	\label{tab3}
\end{table}

\begin{table}
	\caption{The trigger identification and role classification results based on pre-trained RoBERTa-wwm-ext}
	\centering
	\begin{tabular} {p{70pt} p{110pt} p{45pt} p{45pt} p{45pt} p{50pt}} %{llllll} 
		\toprule %\hline
		Tasks	& Models	& Precision	& Recall	& F1 & Loss/epoch \\
		\midrule %\hline
		\multirow[c]{4}{0.6 in}{Trigger}	& BERT & 0.802	& 0.848	& 0.8244	& 0.0634	\\
		& POS-DP-LSTM-BERT	& 0.7965	& 0.8355	& 0.8156	& 0.0731	\\			
		& POS-Embedding-BERT	& 0.9818	& 0.9859	& 0.9838	& 0.0026	\\	
		& DP-GCN-BERT	& 0.9793	& 0.9834	& 0.9814	& 0.0035	\\	
		& POS-DP-GCN-BERT	& 0.981	& 0.9884	& 0.9847	& 0.0034	\\
		\midrule %\hline
		\multirow[c]{4}{0.6 in}{Role}	& BERT & 0.5385	& 0.6569	& 0.5919	& 0.3383	\\
		& POS-DP-LSTM-BERT	& 0.5125	& 0.634	& 0.5668	& 0.4212	\\			
		& POS-Embedding-BERT	& 0.8073	& 0.8621	& 0.8338	& 0.0568	\\	
		& DP-GCN-BERT	& 0.7984	& 0.848	& 0.8225	& 0.0552	\\
		& POS-DP-GCN-BERT	& 0.804	& 0.8696	& 0.8355	& 0.0521		\\
		
		\bottomrule %\hline
	\end{tabular}
	\label{tab4}
\end{table}

From Table \ref{tab3} and Table \ref{tab4}, some findings can be summarized as follows:

Firstly, the integrated POS-DP-GCN-BERT model achieves the best performance in the Trigger subtask. The F1 scores are 99.59\% and 98.47\%, respectively, when using RBT3 and RoBERTa-wwm-ext as benchmark models.

Secondly, in the role classification subtask, the integrated POS-DP-GCN-BERT model achieves the best performance when using RoBERTa-wwm-ext as a benchmark model. The F1 scores are 83.55\%. But the integrated model POS-DP-GCN-BERT (84.41\% F1 score) is inferior to POS-Embedding-BERT (86.68\% F1 score) and DP-GCN-BERT (86.34\% F1 score), which only combine one kind of syntactic feature.

Thirdly, from the above performance, we speculate that POS and DP features are both beneficial for Trigger identification and Role classification. More specifically, the POS features look more important than the DP features. However, this needs more experiments to verify this speculation.

Lastly, more experiments are needed to explore the details of the reasons why a smaller integrated BERT model (RBT3 as the benchmark) performs better than a more complex integrated BERT model (RoBERTa-wwm-ext as the benchmark), e.g., overfitting or underfitting problems.

\section{Conclusion}
This paper proposes an integrated framework to perform Chinese Event Extraction. It considers both the semantic features and syntactic features. The empirical experiments show that the Part of Speech and Dependency Parsing can significantly enhance the event extraction performance.However, there are still some shortcomings. Thus, in the future more datasets and experiments are required to verify some speculations obtained according to the current results.

\bibliographystyle{unsrt}  
\bibliography{eerefbib}

\end{document}